%% file: main.tex
\documentclass[10pt,twocolumn,letterpaper]{article}

\usepackage{cvpr}

\usepackage{times}

\usepackage{multirow}
\usepackage{rotating}
\usepackage{booktabs}
\usepackage{slashbox} 
\usepackage{algpseudocode}
\usepackage{algorithm}

\usepackage{epsfig}
\usepackage{graphicx}
\usepackage{amsmath}
\usepackage{amssymb}
\usepackage{graphicx}
\usepackage{caption}
\usepackage{subcaption}
\usepackage{soul,xcolor}
\usepackage{paralist}
\usepackage{soul}
\usepackage[normalem]{ulem}

\usepackage{mysymbols}

\graphicspath{ {images/} }

\usepackage[pagebackref=true,breaklinks=true,letterpaper=true,colorlinks,bookmarks=false]{hyperref}

\newcommand{\change}[1]{\textcolor{black}{#1}}

\cvprfinalcopy 

\def\cvprPaperID{911} 

\ifcvprfinal\fi
\begin{document}
\setstcolor{blue}
\title{Yin and Yang: Balancing and Answering Binary Visual Questions}


\author{Peng Zhang\thanks{The first two authors contributed equally.} $\,^\dagger$  
  \quad Yash Goyal\footnotemark[1] $\,^\dagger$ \quad Douglas Summers-Stay$^\ddagger$ \quad Dhruv Batra$^\dagger$ \quad Devi Parikh$^\dagger$\\
$^\dagger$Virginia Tech \quad $^\ddagger$Army Research Laboratory\\
{\tt\small $^\dagger$\{zhangp, ygoyal, dbatra, parikh\}@vt.edu} \quad {\tt\small $^\ddagger$douglas.a.summers-stay.civ@mail.mil}}

\maketitle

\input{abstract}
\input{introduction}
\input{related_work}
\input{dataset}

\input{approach}

\input{results_fill}
\input{discussion}

\input{conclusion}
\input{language_prior}
\input{dataset_ap}

\input{qualitative}

\input{negation}
\input{features}
\input{tuple_extraction}
\input{append}

{
\footnotesize
\bibliographystyle{ieee}
\bibliography{references}
}

\end{document}

%% file: abstract.tex
\begin{abstract}


The complex compositional structure of language makes problems at the intersection of vision and language challenging. But language also provides a strong prior that can result in good superficial performance, without the underlying models truly understanding the visual content.  This can hinder progress in pushing state of art in the computer vision aspects of multi-modal AI.

In this paper, we address binary Visual Question Answering (VQA) on abstract scenes. We formulate this problem as \emph{visual verification} of concepts inquired in the questions. 
\change {Specifically, we convert the question to a tuple that 
concisely summarizes the visual concept to be detected in the image. 
If the concept can be found in the image,} the answer to the question is ``yes'', and otherwise ``no''. 
Abstract scenes play two roles (1) They allow us to focus on the high-level semantics of the VQA task as opposed to the low-level recognition problems, and perhaps more importantly, (2) They provide us the modality to \emph{balance} the dataset such that language priors are controlled, and the role of vision is essential. In particular, we collect fine-grained pairs of scenes for every question, such that the answer to the question is ``yes'' for one scene, and ``no'' for the other \emph{for the exact same question}. Indeed, language priors alone do not perform better than chance on our balanced dataset. Moreover, our proposed approach 
matches the performance of a state-of-the-art VQA approach on the unbalanced dataset, and outperforms it on the balanced dataset.
\end{abstract}

%% file: introduction.tex
\section{Introduction}
\label{sec:introduction}
\vspace{-2pt}
Problems at the intersection of vision and language are increasingly drawing more attention. We are witnessing a move beyond the classical ``bucketed'' recognition paradigm (e.g. label every image with categories) to rich compositional tasks involving natural language. 
Some of these problems concerning vision and language have proven surprisingly easy to take on with relatively simple techniques. Consider image captioning, which involves generating a sentence describing a given image \cite{captioning_msr,captioning_xinlei,captioning_berkeley,captioning_baidu_ucla,
captioning_toronto,captioning_stanford,captioning_google}. It is possible to get state of the art results with a relatively coarse understanding of the image by exploiting the statistical biases (inherent in the world and in particular datasets) that are captured in standard language models.

\begin{figure}[t]
\centering
\includegraphics[width=0.45\textwidth]{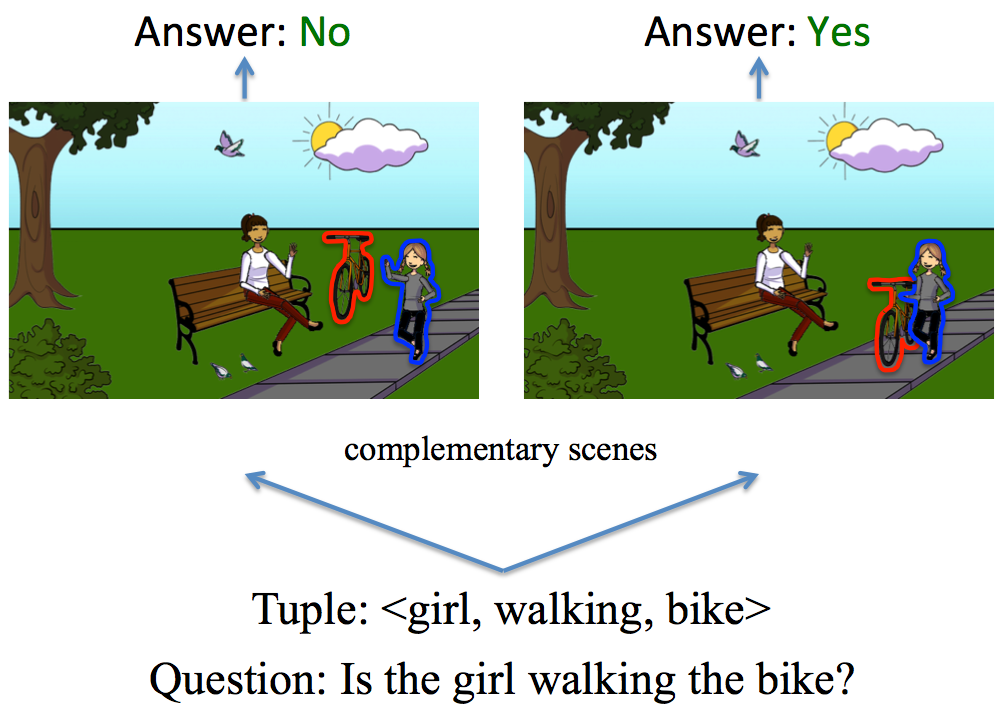}
\vspace{-3pt}
\caption{We address the problem of answering binary questions about images. To eliminate strong language priors that shadow the role of detailed visual understanding in visual question answering (VQA), we use abstract scenes to collect a \emph{balanced} dataset containing pairs of \emph{complementary} scenes: the two scenes have opposite answers to the same question, while being visually as similar as possible. We view the task of answering binary questions as a visual verification task: 
\change{we convert the question into a tuple that concisely summarizes the visual concept, which if present, result in the answer of the question being ``yes'', and otherwise ``no''.} 
Our approach attends to relevant portions of the image when verifying the presence of the visual concept.}
\label{fig:teaser_fig}
\vspace{-10pt}
\end{figure}

For example, giraffes are usually found in grass next to a tree in the MS COCO dataset images \cite{coco}. Because of this, the generic caption ``A giraffe is standing in grass next to a tree'' is applicable to most images containing a giraffe in the dataset. The machine can confidently generate this caption just by recognizing a ``giraffe'', without recognizing ``grass'', or ``tree'', or ``standing'', or ``next to''. In general, captions borrowed from nearest neighbor images result in a surprisingly high performance \cite{Larry_NN_caption}.

A more recent task involving vision and language is Visual Question Answering (VQA). A VQA system takes an image and a free-form natural language question about the image as input (e.g. ``What is the color of the girl's shoes?'', or ``Is the boy jumping?''), and produces a natural language answer as its output (e.g. ``blue'', or ``yes''). Unlike image captioning, answering questions requires the ability to identify specific details in the image (e.g. color of an object, or activity of a person). There are several recently proposed VQA datasets on real images e.g. \cite{VQA, fritz, Malinowski_2015_ICCV, baiduVQA, Ren_2015_NIPS}, as well as on abstract scenes \cite{VQA}. The latter allows research on semantic reasoning without first requiring the development of highly accurate detectors.


Even in this task, however, a simple prior can give the right answer a surprisingly high percentage of the time. 
For example, \change{in the VQA dataset (with images from MS COCO) \cite{VQA},} the most common sport answer ``tennis'' is the correct answer for 41\% of the questions starting with ``What sport is''. Similarly, ``white'' alone is the correct answer for 23\% of the questions starting with ``What color are the''.
Almost half of all questions in the VQA datatset \cite{VQA} can be answered correctly by a neural network that ignores the image completely and uses the question alone, relying on systematic regularities in the kinds of questions that are asked and what answers they tend to have.

This is true even for binary questions, where the answer is either ``yes'' or ``no'', such as ``Is the man asleep?'' or ``Is there a cat in the room?''. One would think that without considering the image evidence, both answers would be equally plausible. Turns out, one can answer 68\%  of binary questions correctly by simply answering ``yes'' to all binary questions. 
Moreover, a language-only neural network can correctly answer more than 78\% of the binary questions, without even looking at the image.

As also discussed in \cite{bias}, such dataset bias effects can give a false impression that a system is making progress towards the goal of understanding images correctly.
Ideally, we want language to pose challenges involving the visual understanding of rich semantics while not allowing the systems to get away with ignoring the visual information. 
Similar to the ideas in \cite{geman}, we propose to unbias the dataset, which would force machine learning algorithms to exploit \emph{image} information in order to improve their scores instead of simply learning to game the test. This involves not only having an equal number of ``yes'' and ``no'' answers on the test as a whole, but also ensuring that \emph{each particular question is unbiased}, so that the system has no reason to believe, without bringing in visual information, that a question should be answered with ``yes'' or ``no.''

In this paper, we focus on binary (yes/no) questions for two reasons. 
First, unlike open-ended questions (Q: ``what is the man playing?'' A: ``tennis''), in binary questions (Q: ``is the man playing tennis?'') all relevant semantic information (including ``tennis'') is available in the question alone. Thus, answering binary questions can be naturally viewed as \emph{visual verification} of concepts inquired in the question 
(``man playing tennis'').
Second, binary questions are easier to evaluate than open-ended questions.

Although our approach of visual verification is applicable to real images (more discussion in \secref{sec:discussion}), we choose to use abstract images \cite{VQA, Antol2014, ZitnickICCV2013, ZitnickCVPR2013, VedantamPAMI2015} as a test bed because abstract scene images allow us to focus on high-level semantic reasoning. They also allow us to balance the dataset by making changes to the images, something that would be difficult or impossible with real images. 


Our main contributions are as follows:
(1) We balance the existing abstract binary VQA dataset \cite{VQA} by creating complementary scenes so that all questions\footnote{nearly all. About 6\% of test questions do not lend themselves to this modification. See \secref{sec:dataset} for details.} have an answer of ``yes'' for one scene \emph{and} an answer of ``no'' for another closely related scene. We show that a language-only approach performs significantly worse on this balanced dataset. 
(2) We propose an approach that summarizes the content of the question in a \change{tuple form which
concisely describes the 
visual concept 
whose existence is to
be verified in the scene. 
We answer the question by verifying if the tuple is depicted in the scene or not (See \figref{fig:teaser_fig}).}
We present results when training and testing on the balanced and unbalanced datasets. 



%% file: related_work.tex
\section{Related work}
\label{sec:relatedwork}

\textbf{Visual question answering.} 
Recent work has proposed several datasets and methods to promote research on the task of visual question answering \cite{geman, vizwiz, SongChun_video_queries, fritz, VQA, Malinowski_2015_ICCV, baiduVQA, Ren_2015_NIPS}, ranging from constrained settings \cite{geman, fritz, Ren_2015_NIPS} to free-form natural language questions and answers \cite{vizwiz, SongChun_video_queries, VQA, Malinowski_2015_ICCV, baiduVQA}. For example,
\cite{geman} proposes a system to generate binary questions from templates using a fixed vocabulary of objects, attributes, and relationships between objects. 
\cite{SongChun_video_queries} has studied joint parsing of videos and corresponding text to answer queries about videos. 
\cite{fritz} studied VQA with synthetic (templated) and human-generated questions, both with the restriction of answers being limited to 16 colors and 894 object categories or sets of categories.
A number of recent papers~\cite{VQA, baiduVQA, Malinowski_2015_ICCV, Ren_2015_NIPS} proposed neural network models for VQA composing LSTMs (for questions) and CNNs (for images). 
\cite{VQA} introduced a large-scale dataset for free-form and open-ended VQA, along with several natural VQA models.
\cite{vizwiz} uses crowdsourced workers to answer questions about visual content asked by visually-impaired users.


\textbf{Data augmentation.}
Classical data augmentation techniques (such as mirroring, cropping) have been widely used in past few years \cite{Howard_DataAug, MSR_DataAug} to provide high capacity models additional data to learn from.
These transformations are designed to not change the label distribution in the training data. 
In this work, we ``augment'' our dataset to explicitly change the label distribution. 
We use human subjects to collect additional scenes such that \emph{every question} in our dataset has equal number of `yes' and `no' answers (to the extent possible). In that sense, our approach can be viewed as semantic data augmentation.
Several classification datasets, such as ImageNet \cite{ImageNet} try to be balanced. But this is infeasible for the VQA task on real images because of the heavy-tail of concepts captured by language.
This motivates our use of abstract scenes.

\textbf{Visual abstraction + language.} 
A number of works have used abstract scenes to focus on high-level semantics and study its connection with other modalities such as language \cite{Lin_2015_CVPR, abstract_captioning, ZitnickCVPR2013, VedantamPAMI2015, ZitnickICCV2013, Antol2014, clipart_dynamics, Vendantam_2015_ICCV}, 
including automatically describing abstract scenes \cite{abstract_captioning}, 
generating abstract scenes that depict a description \cite{ZitnickICCV2013}, 
capturing common sense \cite{clipart_dynamics, Lin_2015_CVPR, Vendantam_2015_ICCV}, 
learning models of fine-grained interactions between people \cite{Antol2014}, 
and learning the semantic importance of visual features \cite{ZitnickCVPR2013, VedantamPAMI2015}.
Some of these works have also taken advantage of visual abstraction to ``control" the distribution of data, for example, 
\cite{Antol2014} collects equal number of examples for each verb/preposition combinations, and 
\cite{ZitnickCVPR2013} have multiple scenes that depict the exact same sentence/caption. 
Similarly, we balance the dataset by making sure that each question in the dataset has a scene for ``yes'' and another scene for ``no'' to the extent possible.

\textbf{Visual verification.}
\cite{Sadeghi_2015_CVPR, Vendantam_2015_ICCV} reason about the plausibility of commonsense assertions (men, ride, elephants) by gathering visual evidence for them in real images~\cite{Sadeghi_2015_CVPR} and abstract scenes~\cite{Vendantam_2015_ICCV}.
In contrast, we focus on visually-grounded image-specific questions like ``Is the man in the picture riding an elephant?".
\cite{ZitnickICCV2013} also reasons about relations between two objects, and maps these relations to visual features. They take as input a description and automatically generate a scene that is compatible with all tuples in the description and is a plausible scene. In our case, we have a \change{single tuple (summary of the question) and we want to verify if 
it}
exists in a given image or not, for the goal of answering a free form ``yes/no'' question about the image. 

\change{\textbf{Visual attention} involves searching and attending to relevant image regions. 
\cite{karpathy14, xu_caption} uses alignment/attention for image caption generation. Input is just an image, and they try to describe the entire image and local regions with phrases and sentences. We address a different problem: visual question answering. We are given an image \emph{and text} (a question) as input. We want to align parts of the question to regions in the image so as to extract detailed visual features of the regions of the image being referred to in the text. 
}






%% file: dataset.tex
\section{Datasets}
\label{sec:dataset}

We first describe the VQA dataset for abstract scenes collected by \cite{VQA}. We then describe how we balance this dataset by collecting more scenes.

\subsection{VQA dataset on abstract scenes}
\label{subsec:VQAdataset}

\textbf{Abstract library.}
The clipart library contains 20 ``paperdoll'' human models \cite{Antol2014} spanning genders, races,
and ages with 8 different expressions. The limbs are adjustable to allow
for continuous pose variations. 
In addition to humans, the library contains 99 objects and 31 animals in various poses. 
The library contains two different scene types -- ``indoor'' scenes, containing only indoor objects, e.g. desk, table, etc., and ``outdoor'' scenes, which contain outdoor objects, e.g. pond, tree, etc.
The two different scene types are indicated by different background in the scenes.

\textbf{VQA abstract dataset}
consists of 50K abstract scenes,
with 3 questions for each scene,
with train/val/test splits of 20K/10K/20K scenes respectively.
This results in total 60K train, 30K validation and 60K test questions.
Each question has 10 human-provided ground-truth answers.
Questions are categorized into 3 types -- `yes/no', `number', and `other'.
In this paper, we focus on `yes/no' questions, which gives us a dataset of 36,717 questions- 24,396 train and 12,321 val questions.
Since test annotations are not publicly available, it is not possible to find the number of `yes/no' type questions in test set. 
\change{
We use the binary val questions as our unbalanced test set, a random subset of 2,439 training questions as our unbalanced validation set, and rest of the training questions as our unbalanced train set.}

\begin{figure}[t]
\centering
\includegraphics[width=1\linewidth]{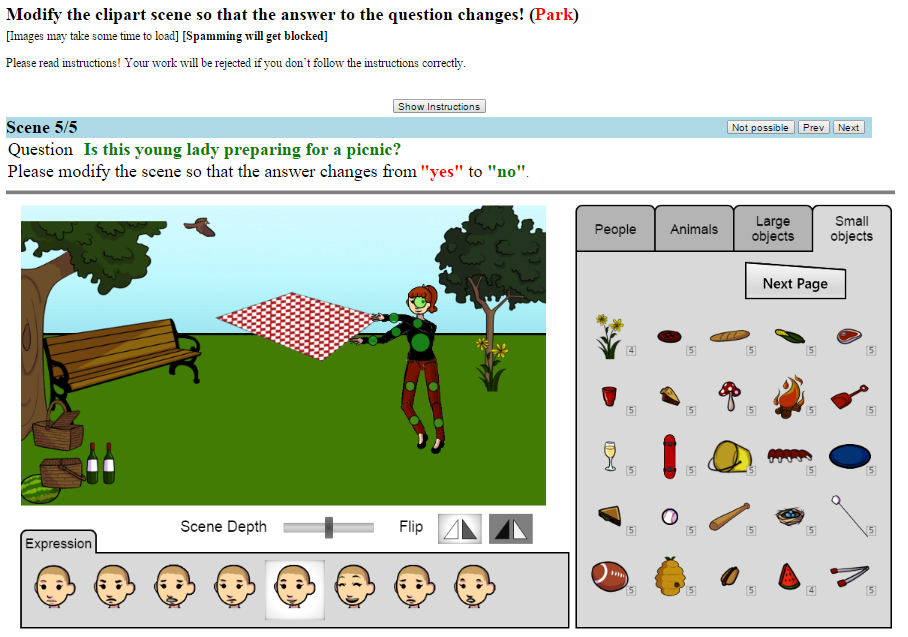}
\vspace{-12pt}
\caption{A snapshot of our Amazon Mechanical Turk (AMT) interface to collect complementary scenes.}
\vspace{-10pt}
\label{fig:interface}
\end{figure} 

\subsection{Balancing abstract binary VQA dataset}
\label{subsec:balanced_dataset}

We balance the abstract VQA dataset by posing a counterfactual task -- given an abstract scene and a binary question, 
what would the scene have looked like \emph{if the answer to the binary question was different}? While posing such counterfactual questions and obtaining corresponding scenes is nearly impossible in real images, abstract scenes allow us to perform such reasoning. 

We conducted the following Mechanical Turk study -- given an abstract scene, and an associated question from the VQA dataset, we ask subjects to \emph{modify the clipart scene such that the answer changes} from `yes' to `no' (or `no' to `yes'). 
For example, for the question ``Is a cloud covering the sun?'', a worker can move the `sun' into open space in the scene to change the answer from `yes' to `no'.
A snapshot of the interface is shown in \figref{fig:interface}. 

We ask the workers to modify the scene as little as possible. 
We encourage minimal changes because
these complementary scenes can be thought of as hard-negatives/positives to learn subtle differences in the visual signal that are relevant to answering questions. 
This signal can be used as additional supervision for training models such as \cite{Zaidan,Donahue,AmarECCV2012,BiswasCVPR2013} that can leverage explanations provided by the annotator in addition to labels.
Our complementary scenes can also be thought of as analogous to good pedagogical techniques where a learner is taught concepts by changing one thing at a time via contrasting (e.g., one fish vs. two fish, red ball vs. blue ball, etc.). 
Full instructions on our interface can be found in supp. material.

Note that there are some (scene, question) pairs that do not lend themselves to easy creation of complementary scenes with the existing clipart library.
For instance, if the question is ``Is it raining?'', and the answer needs to be changed from `no' to `yes', it is not possible to create `rain' in the current clipart library. 
Fortunately, these scenes make up a small minority of the dataset (e.g., 6\% of the test set). 




\change{
To keep the balanced train and test set comparable to unbalanced ones in terms of size, we collect complementary scenes for $\sim$half of the respective splits -- 11,760 from train and 6,000 from test set.  
Since Turkers indicated that 2,137 scenes could not be modified to change the answer because of limited clipart library, we do not have complementary scenes for them. In total, we have 10,295 complementary scenes for the train set 
and 5,328 complementary scenes for test, resulting in balanced train set containing 22,055 samples and balanced test set containing 11,328 samples. We further split a balanced set of 2,202 samples from balanced train set for validation purposes.}
Examples from our balanced dataset are shown in  \figref{fig:teaser_fig} and \figref{fig:qualitative_example}.

We use the publicly released VQA evaluation script in our experiments. The evaluation metric uses 10 ground-truth answers for each question to compute performance. To be consistent with the VQA dataset, we collected 10 answers 
from human subjects using AMT for all complementary scenes in the balanced test set.

We compare the degree of balance in our unbalanced and balanced datasets.
We find that 92.65\% of the (scene, question) pairs in the unbalanced test set do not have a corresponding complementary scene (where the answer to the same question is the opposite). Only 20.48\% of our balanced test set does not have corresponding complementary scenes. Note that our dataset is not 100\% balanced either because there are some scenes which could not be modified to flip the answers to the questions (5.93\%) or because the most common answer out of 10 human annotated answers for some questions does not match with the intended answer of the person creating the complementary scene (14.55\%) \change{either due to inter-human
disagreement, or if the worker did not succeed in creating a good scene.}

%% file: approach.tex
\section{Approach}
\label{sec:approach}
We present an overview of our approach before describing each step in detail in the following subsections.
To answer binary questions about images, we propose a two-step approach: (1) Language Parsing: where the question is parsed into a tuple, and (2) Visual Verification: 
where we verify whether that tuple is present in the image or not.


Our language parsing step summarizes a binary question into a tuple of the form $<$P, R, S$>$, where P refers to primary object, R to relation and S to secondary object, 
e.g. for a binary question ``Is there a cat in the room?'', our goal is to extract a tuple of the form: $<$cat, in, room$>$. 
Tuples need not have all the arguments present. For instance, ``Is the dog asleep'' $\rightarrow$ $<$dog, asleep, $>$,
The primary argument P is always present.
Since we only focus on binary questions, this extracted tuple captures the 
\change{entire visual concept to be verified in the image.} If the concept is depicted in the image, the answer is ``yes'', otherwise the answer is ``no''.

Once we extract $<$P, R, S$>$ tuples from questions (details in \secref{subsec:tuple_extraction}), we align the P and S arguments to objects in the image (\secref{subsec:aligning_text_image}). We then extract text and image features (\secref{subsec:features}), and finally learn a model to reason about the consistency of the 
\change{tuple} 
with the image (\secref{subsec:visual_verification}). 



\subsection{Tuple extraction}
\label{subsec:tuple_extraction}
In this section, we describe how we extract $<$P, R, S$>$ tuples from raw questions. 
Existing NLP work such as~\cite{Fader11} has studied this problem, however, these approaches are catered towards statements, and are not directly applicable to questions.
We only give an overview of our method, more details can be found in supp. material.

\textbf{Parsing:} We use the Stanford parser to parse the question. Each word is assigned an entity, e.g. nominal subject (``nsubj''), direct object (``dobj''), etc.
We remove all characters other than letters and digits before parsing. 

\textbf{Summarizing:} As an intermediate step, we first convert a question into a ``summary'', before converting that into a tuple. 
First, we remove a set of ``stop words'' such as determiners (``some'', ``the'', etc.) and auxillary verbs (``is'', ``do'', etc.). 
Our full list of stop words is provided in supp. material.
Next, following common NLP practice, we remove all words before a nominal subject (``nsubj'') or a passive nominal subject (``nsubjpass''). 
For example, ``Is the woman on couch petting the dog?'' is parsed as ``Is(\emph{aux}) the(\emph{det}) woman(\emph{nsubj}) on(\emph{case}) couch(\emph{nmod}) petting(\emph{root}) the(\emph{det}) dog(\emph{dobj})?''. The summary of this question can be expressed as (woman, on, couch, petting, dog). 

\textbf{Extracting tuple:} Now that we have extracted a summary of each question, next we split it into PRS arguments. 
Ideally, we would like P and S to be noun phrases (``woman on couch'', ``dog'') and the relation R to be a verb phrase (``petting'') or a preposition (``in'') when the verb is a form of ``to be''. For example, $<$dog, in, room$>$, or $<$woman on couch, petting, dog$>$. Thus, we apply the Hunpos Part of Speech (POS) tagger~\cite{hunpos} to assign words to appropriate arguments of the tuple. 
See supp. material for details.

\subsection{Aligning objects to primary (P) and secondary (S) arguments}
\label{subsec:aligning_text_image}
In order to extract visual features that describe the objects in the scene being referred to by P and S, we need to align each of them with the image. 

We extract PRS tuples from all binary questions in the training data. Among the three arguments, P and S contain noun phrases. 
To determine which objects are being referred to by the P and S arguments, we follow the idea in \cite{ZitnickICCV2013} and  compute the mutual information\footnote{We compute MI separately for indoor and outdoor scenes. More details about scene types can be found in \secref{subsec:VQAdataset}.} between word occurrence
(e.g. ``dog''), and object occurrence (e.g. clipart piece \#32). 
We only consider P and S arguments that occur at least twice in the training set. 
At test time, given an image and a PRS tuple corresponding to a binary question, the object in the image with the highest mutual information with P is considered to be referred by the primary object, and similarly for S. 
If there is more than one instance of the object category in the image, we assign P/S to a random instance.
Note that for some questions with ground-truth answer `no', it is possible that P or S actually refers to an object that is not present in the image (e.g. Question: ``Is there a cat in the image?'' Answer: ``no''). In such cases, some other object from images (say clipart \#23, which is a table) will be aligned with P/S. However, since the category label (`table') of the aligned object is a feature, the model can learn to handle such cases, i.e., learn that when the question mentions `cat' and the aligned clipart object category is `table', the answer should be `no'.

\begin{figure}
	\includegraphics[width=0.5\textwidth]{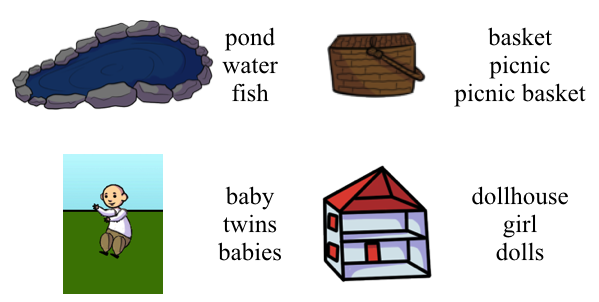}
    \vspace{-12pt}
    \caption{Most plausible words for an object determined using mutual information.}
    \label{fig:mutual_info}
    \vspace{-15pt}
\end{figure}

We found that this simple mutual information based alignment approach does surprisingly well. 
This was also found in \cite{ZitnickICCV2013}.
Fig.~\ref{fig:mutual_info} shows examples of clipart objects and three words/phrases that have the highest mutual information. 

\subsection{Visual verification}
\label{subsec:visual_verification}

We have extracted PRS tuples and aligned PS to the clipart objects in the image, we can now compute a score indicating the strength of visual evidence for the concept inquired in the question.
Our scoring function measures compatibility between text and image features (described in ~\secref{subsec:features}).

\change{
Our model is an ensemble of two similar models-- Q-model and Tuple-model, whose common architecture is inspired from a recently proposed VQA approach~\cite{VQA}. Specifically, each model takes two inputs (image and question), each along a different branch. The two models (Q-model and Tuple-model) use the same image features, but different language features. Q-model encodes the sequential nature of the question by feeding it to an LSTM and using its 256-dim hidden representation as a language embedding, while Tuple-model focuses on the important words in the question and uses concatenation of word2vec \cite{word2vec} embeddings (300-dim) of P, R and S as the language features.
If P, R or S consist of more than one word, we use the average of the corresponding word2vec embeddings. 
This 900-dimensional feature vector is passed through a fully-connected layer followed by a \emph{tanh} non-linearity layer to create a dense 256-dim language embedding. }


The image is represented by rich semantic features, described in \secref{subsec:features}. Our binary VQA model converts these image features into 256-dim with an inner-product layer, followed by a \emph{tanh} layer. This inner-product layer learns to map visual features onto the space of text features. 

Now that both image and text features are in a common space, they are point-wise multiplied resulting in a 256-dim \emph{fused} \change{language+image} representation. This fused vector is then passed through two more fully-connected layers in a Multi-Layered Perceptron (MLP), which finally outputs a 2-way softmax score for the answers `yes' and `no'. 
These predictions from the Q-model and Tuple-model are multiplied to obtain the final prediction. Both the models are learned separately and end-to-end (including LSTM) with a cross-enptropy loss.
Our implementation uses Keras~\cite{keras}. 
Learning is performed via SGD with a batch-size of 32, 
dropout probability 0.5, and the model is trained 
\change{till the validation loss plateaus.}

\change{At test time, given the question and image features, we can perform visual verification simply by performing forward pass through our network.} 

\subsection{Visual Features}
\label{subsec:features}

We use the same features as \cite{Lin_2015_CVPR} for our
approach.
These visual features describe the objects in the image that are being referred to by the P and S arguments,
their interactions, and the context of the scene within which these objects are present. In particular, the feature vector for each scene has 1432 dimensions, which are composed of 563 dimensions for each primary object and secondary object, encoding object category (e.g., cat vs. dog vs. tree), instance (e.g., which particular tree), flip (i.e., facing left or right), absolute location modeled via GMMs, pose (for humans and animals), expression, age, gender and skin color (for humans), 48 dimensions for relative location between primary and secondary objects (modeled via GMMs), and 258 dimensions encoding which other object categories and instances are present in the scene around P and S.



%% file: results_fill.tex
\section{Experiments}
\label{sec:results}

\subsection{Baselines}
\label{subsec:baselines}
We compare our model with several strong baselines including language-only models as well as a state-of-the-art VQA method.

\textbf{Prior:} 
Predicting the most common answer in the training set, for all test questions. The most common answer is ``yes'' in the unbalanced set, and ``no'' in the balanced set.

\textbf{Blind-Q+Tuple:} 
\change{A language-only baseline which has a similar architecture as our approach except that each model only accepts language input and does not utilize any visual information.
Comparing our approach to Blind-Q+Tuple quantifies to what extent our model has succeeded in leveraging the image to answer questions correctly.}


\textbf{SOTA Q+Tuple+H-IMG:} \change{This VQA model has a similar architecture as our approach, except that it uses holistic image features (H-IMG) that describe the entire scene layout,
instead of focusing on specific regions in the scene as determined by P and S. }
This model is analogous to the state-of-the-art models presented in \cite{VQA, Malinowski_2015_ICCV,  Ren_2015_NIPS, baiduVQA}, except applied to abstract scenes.

These holistic features include a bag-of-words for clipart objects occurrence (150-dim), human expressions (8-dim), and human poses (7-dim). 
The 7 human poses refer to 7 clusters obtained by clustering all the human pose vectors (concatenation of (x, y) locations and global angles of all 15 deformable parts of human body) in the training set. 
We extract these 165-dim holistic features for the complete scene and for four quadrants, and concatenate them together to create a 825-dim vector.
These holistic image features are similar to decaf features for real images, which are good at capturing what is present where, but (1) do not attend to different parts of the image based on the questions, and (2) may not be capturing intricate interactions between objects.

Comparing our model to SOTA Q+Tuple+H-IMG quantifies the improvement in performance by attending to specific regions in the image as dictated by the question being asked, and explicitly capturing the interactions between the relevant objects in the scene. In other words, we quantify the improvement in performance obtained by pushing for a deeper understanding of the image than generic global image descriptors.
Thus, we name our model Q+Tuple+A-IMG, where A is for attention. 


\subsection{Evaluation on the original (unbalanced) dataset}
\label{subsec:imbalanced_set}
In this subsection, we train all models on the train splits of both the unbalanced and balanced datasets, and test on our unbalanced test set.  
The results are shown in Table~\ref{tab:imbalanced}.


\begin{table}[h]
\setlength{\tabcolsep}{10pt}
{\small
\begin{center}
\begin{tabular}{@{}lcc@{}}
\toprule
 & \multicolumn{2}{@{}c@{}}{Training set} \\
\toprule
 & Unbalanced & Balanced \\
\midrule
  Prior (``yes'') &  68.67 & 68.67\\
  Blind-Q+Tuple &  78.90 & 60.80\\ 
\emph{SOTA} Q+Tuple+H-IMG & 78.49 & 69.19\\
  \emph{Ours} Q+Tuple+A-IMG & \textbf{79.20} & \textbf{72.80}\\
\bottomrule
\end{tabular}
\end{center}
}
\vspace{-14pt}
\caption {Evaluation on unbalanced test set. All accuracies are calculated using the VQA \cite{VQA} evaluation metric.}
\label{tab:imbalanced}
\end{table}
\vspace{-7pt}
We draw the following key inferences:

\textbf{Vision helps.} We observe that models that utilize visual information tend to perform better than ``blind'' model when trained on the balanced dataset. This is because the lack of strong language priors in the balanced dataset forces the models to focus on the visual understanding.



\textbf{Attending to specific regions is important.} When trained on the balanced set where visual understanding is critical, our proposed model Q+Tuple+A-IMG, which focuses only on a specific region in the scene, outperforms all the baselines by a large margin. 

\textbf{Bias is exploited.} As expected, the performance of all models trained on unbalanced dataset is better than the balanced dataset, because these models learn the language biases while training on unbalanced dataset, which are also present in the unbalanced test set.






\begin{figure*}[t]
\centering
   \includegraphics[width=\textwidth]{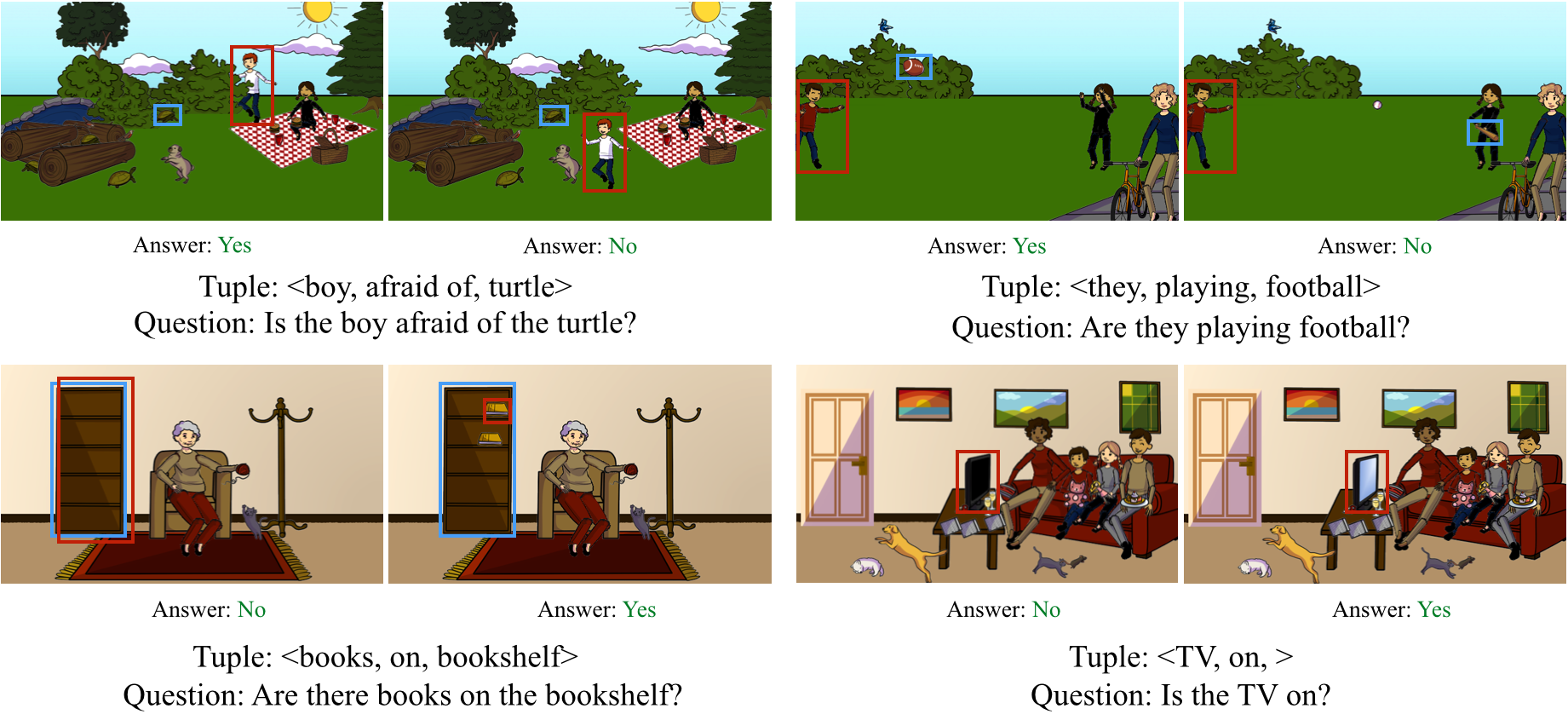}
	\vspace{-16pt}
   \caption{Qualitative results of our approach. We show input questions, complementary scenes that are subtle (semantic) perturbations of each other, along with tuples extracted by our approach, and objects in the scenes that our model chooses to attend to while answering the question. Primary object is shown in red and secondary object is in blue. }
 
\label{fig:qualitative_example}
\vspace{-12pt}
\end{figure*}

\subsection{Evaluation on the balanced dataset}
\label{subsec:balanced_set}

We also evaluate all models trained on the train splits of both the unbalanced and balanced datasets, by testing on the balanced test set. The results are summarized in Table~\ref{tab:balanced}. 


\begin{table}[h]
\setlength{\tabcolsep}{10pt}
{\small
\begin{center}
\begin{tabular}{@{}lcc@{}}
\toprule
 & \multicolumn{2}{@{}c@{}}{Training set} \\
\toprule
 & Unbalanced & Balanced \\
\midrule
  Prior (``no'') &  63.85 & 63.85\\
  Blind-Q+Tuple &  65.98 & 63.33\\ 
  \emph{SOTA} Q+Tuple+H-IMG & 65.89 & 71.03\\
  \emph{Ours} Q+Tuple+A-IMG & \textbf{68.08} & \textbf{74.65}\\
\bottomrule
\end{tabular}
\end{center}
}
\vspace{-14pt}
\caption {Evaluation on balanced test set. All accuracies are calculated using the VQA \cite{VQA} evaluation metric.}
\label{tab:balanced}
\vspace{-12pt}
\end{table}
Here are the observations from this experiment:

\textbf{Training on balanced is better.} It is clear from Table~\ref{tab:balanced} that both language+vision models trained on balanced data perform better than the models trained on unbalanced data. 
This may be because the models trained on balanced data \emph{have} to learn to extract visual information to answer the question correctly, since they are no longer able to exploit language biases in the training set. Where as models trained on the unbalanced set are blindsided into learning strong language priors, which are then not available at test time.

\textbf{Blind models perform close to chance.} As expected, when trained on unbalanced dataset, the ``blind'' model's performance is significantly lower on the balanced dataset (66\%) than on unbalanced (79\%). 
Note that the accuracy is higher than 50\% because this is not binary classification accuracy but the VQA accuracy~\cite{VQA}, which provides partial credit when there is inter-human disagreement in the ground-truth answers.

\textbf{Attention helps.} When trained on balanced dataset (where language biases are absent), our model Q+Tuple+A-IMG is able to outperform all baselines by a significant margin.
Specifically, our model gives improvement in performance relative to the state-of-the-art VQA model from~\cite{VQA} (Q+Tuple+H-IMG), showing that attending to relevant regions and describing them in detail helps, as also seen in \secref{subsec:imbalanced_set}.

\textbf{Role of balancing.} We see clear improvements by reasoning about vision in addition to language. Note that in addition to the lack of language bias, the visual reasoning is also harder on the balanced dataset because now there are pairs of scenes with fine-grained differences but with opposite answers to the same question. So the model really needs to understand the subtle details of the scene to answer questions correctly. Clearly, there is a lot of room for improvement and we hope our balanced dataset will encourage more future work on detailed understanding of visual semantics towards the goal of accurately answering questions about images.


\textbf{Classifying a pair of complementary scenes.}
We experiment with an even harder setting -- 
a test point consists of a pair of complementary scenes and the associated question. Recall, that by construction, the answer to the question is ``yes'' for one image in the pair, and ``no'' for the other.
This test point is considered to be correct only when the model is able to predict both its answers correctly. 

Since language-only models only utilize the textual information in the question ignoring the image, and therefore, predict the same answer for both scenes, their accuracy is zero in this setting\footnote{Note that to create this pair-level test set, we only consider those pairs where the answers were opposites. We removed all scenes that workers were unable to create complementary scenes for due to a finite clipart library, as well as those scenes for which the majority answer from 10 workers did not agree with the intended answer of the creator of the scene.}. The results of the baselines and our model, trained on balanced and unbalanced datasets, are shown in Table~\ref{tab:pair_task}. We observe that our model trained on the balanced dataset performs the best. And again, our model that focuses on relevant regions in the image to answer the question outperforms the state-of-the-art approach of \cite{VQA} (Q+Tuple+H-IMG) that does not model attention.


\begin{table}[h]
\setlength{\tabcolsep}{10pt}
{\small
\begin{center}
\begin{tabular}{@{}lcc@{}}
\toprule
 & \multicolumn{2}{@{}c@{}}{Training set} \\
\toprule
 & Unbalanced & Balanced \\
\midrule
  Blind-Q+Tuple & 0 & 0 \\
  Q+Tuple+H-IMG &  03.20 & 23.13\\
  Q+Tuple+A-IMG &  \textbf{09.84} & \textbf{34.73}\\ 
\bottomrule
\end{tabular}
\end{center}
}
\vspace{-14pt}
\caption {Classifying a pair of complementary scenes. All accuracies are percentage of test pairs that have been predicted correctly.}
\label{tab:pair_task}
\end{table}
\vspace{-7pt}
\subsection{Analysis}
\label{subsec:analysis}
\change{
Our work involves three steps: tuple extraction, tuple and object alignment, and question answering. 
We conduct analyses of these three stages to determine the importance of each of the three stages.
We manually inspected a random subset of questions, and found the tuple extraction to be accurate 86.3\% of the time.
Given perfect tuple extraction, the alignment step is correct 95\% of the time.
Given perfect tuple extraction and alignment, our approach achieves VQA accuracy of 81.06\% as compared to 79.2\% with imperfect tuple extraction and alignment. Thus, $\sim$2\% in VQA accuracy is lost due to imperfect tuple extraction and alignment.
}

\subsection{Ablation Study}
\label{subsec:analysis}

We conducted an ablation study to analyze the importance of the two kinds of language features-- LSTM for question vs. word2vec for tuple. For ``blind'' (language only) models trained and tested on unbalanced datasets, we found that the combination (Q+Tuple) performs better than each of the individual methods. Specifically, Q+Tuple achieves a VQA accuracy of 78.9\% as compared to 77.87\% (Q-only) and 77.54\% (Tuple-only). 

\subsection{Qualitative results}
\label{subsec:qualitative}

\figref{fig:qualitative_example} shows qualitative results for our approach. We show a question and two complementary scenes with opposite answers. We find that even though pairs of scenes with opposite ground truth answers to the same questions are visually similar, our model successfully predicts the correct answers for both scenes. Further, we see that our model has learned to attend to the regions of the scene that seem to correspond to the regions that are most relevant to answering the question at hand. The ability to (correctly) predict different answers to scenes that are subtle (semantic) perturbations of each other demonstrates visual understanding.

%% file: discussion.tex
\section{Discussion}
\label{sec:discussion}

\change{
The idea of balancing a dataset can be generalized to real images. For instance, we can ask MTurk workers to find images with different answers for a given question. The advantage with clipart is that it lets us make the complementary scenes very fine-grained forcing the models to learn subtle differences in visual information. The differences in complementary real images will be coarser and therefore easier for visual models. Overall, there is a trade-off between clipart and real images. Clipart is easier (trivial) for low-level recognition tasks, but is more difficult balanced dataset because it can introduce fine-grained semantic differences. Real is more difficult for low-level recognition tasks, but may be an easier balanced dataset because it will have coarse semantic differences.}

%% file: conclusion.tex
\section{Conclusion}
\label{sec:conclusion}

In this paper, we take a step towards the AI-complete task of Visual Question Answering.
Specifically, we tackle the problem of answering binary questions about images. 
We balance the existing abstract binary VQA dataset
by augmenting the dataset with complementary scenes, so that nearly all questions in the balanced dataset have an answer ``yes'' for one scene and an answer ``no'' for another closely related scene.
For an approach to perform well on this balanced dataset, it \emph{must} understand the image.
We will make our balanced dataset publicly available. 

We propose an approach that extracts a concise summary of the question in a tuple form, identifies the region in the scene it should focus on, and \change{verifies the existence of the visual concept described in the question tuple to answer the question}.
Our approach outperforms the language prior baseline and a state-of-the-art VQA approach by a large margin on the balanced dataset. We also present qualitative results showing that our approach attends to relevant parts of the scene in order to answer the question.


\textbf{Acknowledgements.} 
This work was supported in part by the The Paul G. Allen Family Foundation via an award to D.P., ICTAS at Virginia Tech via awards to D.B. and D.P., Google Faculty Research Awards to D.P. and D.B., the Army Research Lab via awards to D.P. and D.B., the National Science Foundation CAREER award to D.B., the Army Research Office YIP Award to D.B., and an Office of Naval Research grant to D.B. The authors would like to thank Larry Zitnick from Facebook AI Research, Lucy Vanderwende from Microsoft Research and Claire Bonial from Army Research Lab for useful discussions.

%% file: language_prior.tex
\appendix
\section*{Appendix}
\section{Role of language priors}
\label{language_prior}
\begin{figure*}
\centering
   \includegraphics[width=\textwidth]{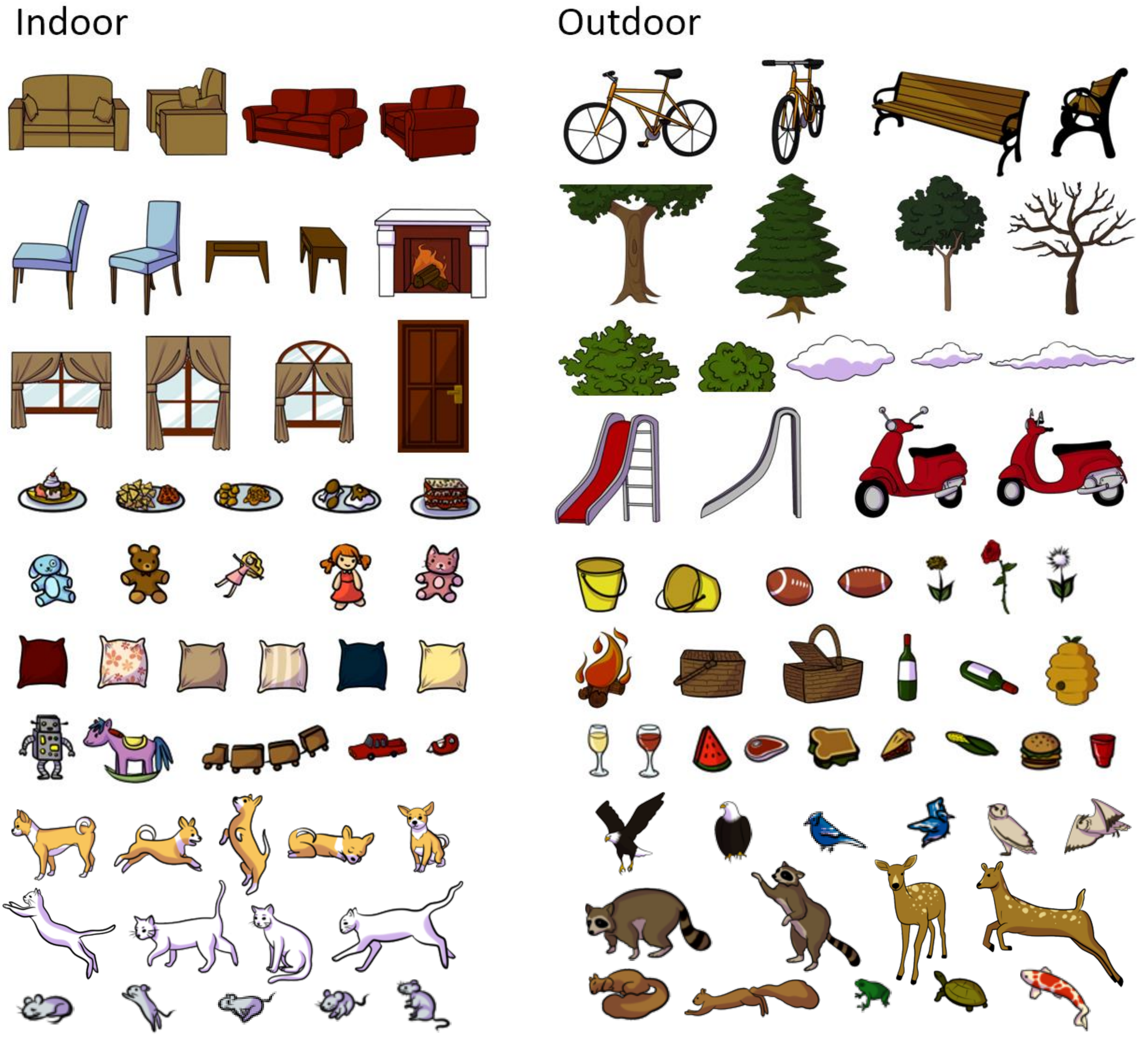}
   \caption{A subset of the clipart objects present in the abstract scenes library.}
\label{fig:clipart_objects}
\end{figure*}

\begin{figure}
\centering
   \includegraphics[width=0.5\textwidth]{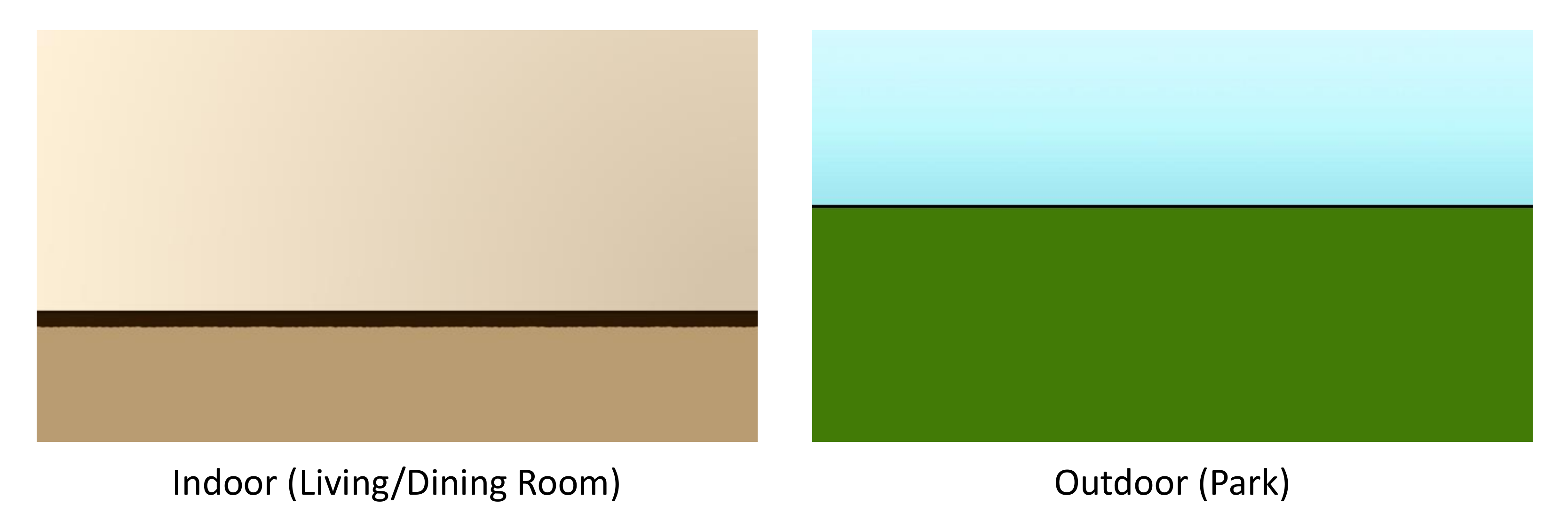}
   \caption{Backgrounds in indoor (left) and outdoor (right) scenes.}
\label{fig:background}
\end{figure}

\begin{figure*}
\centering
   \includegraphics[width=\textwidth]{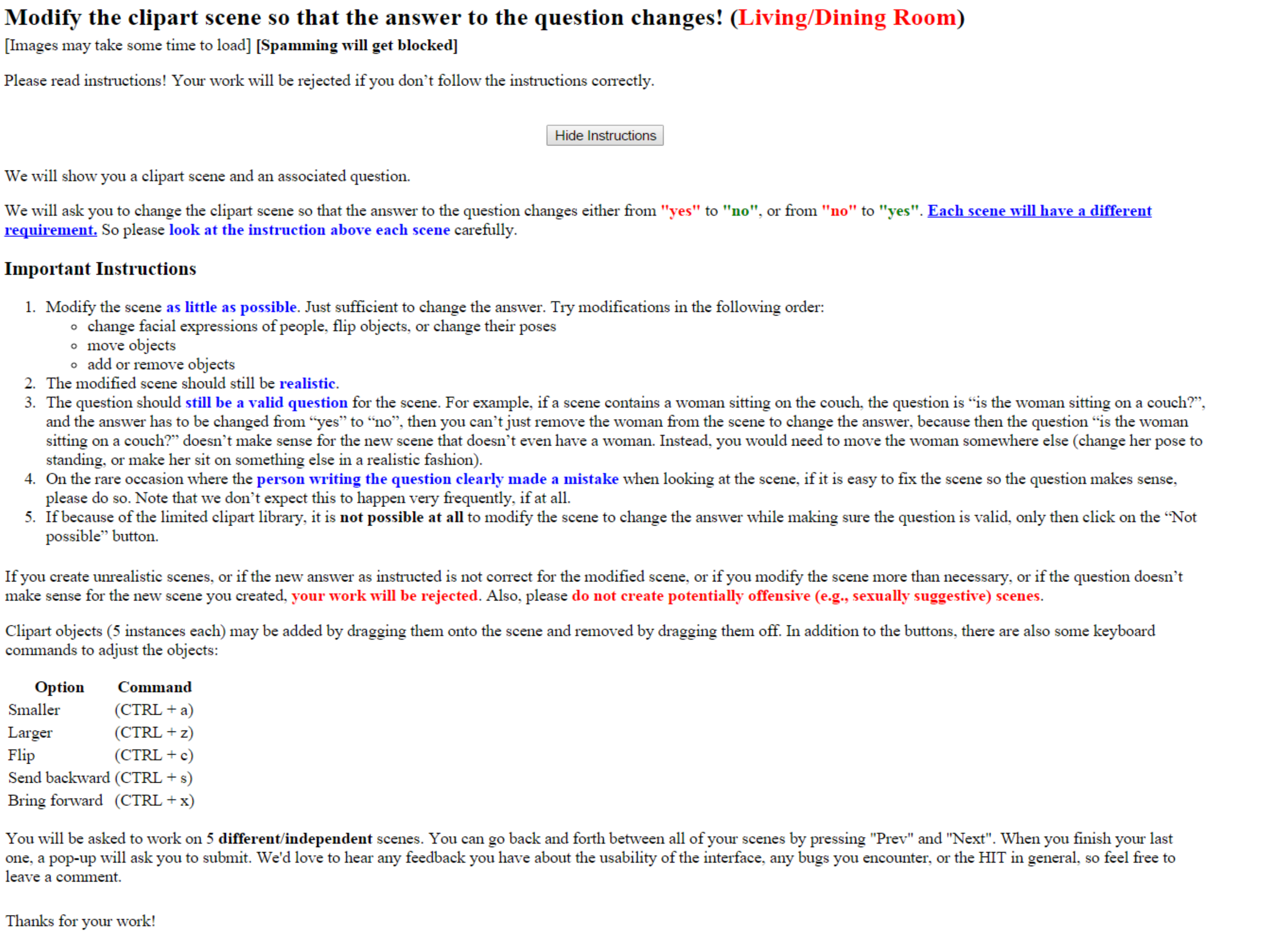}
   \caption{The instructions given to AMT workers while collecting complementary scenes.}
\label{fig:instructions}
\end{figure*}

\begin{figure*}
\centering
   \includegraphics[width=\textwidth]{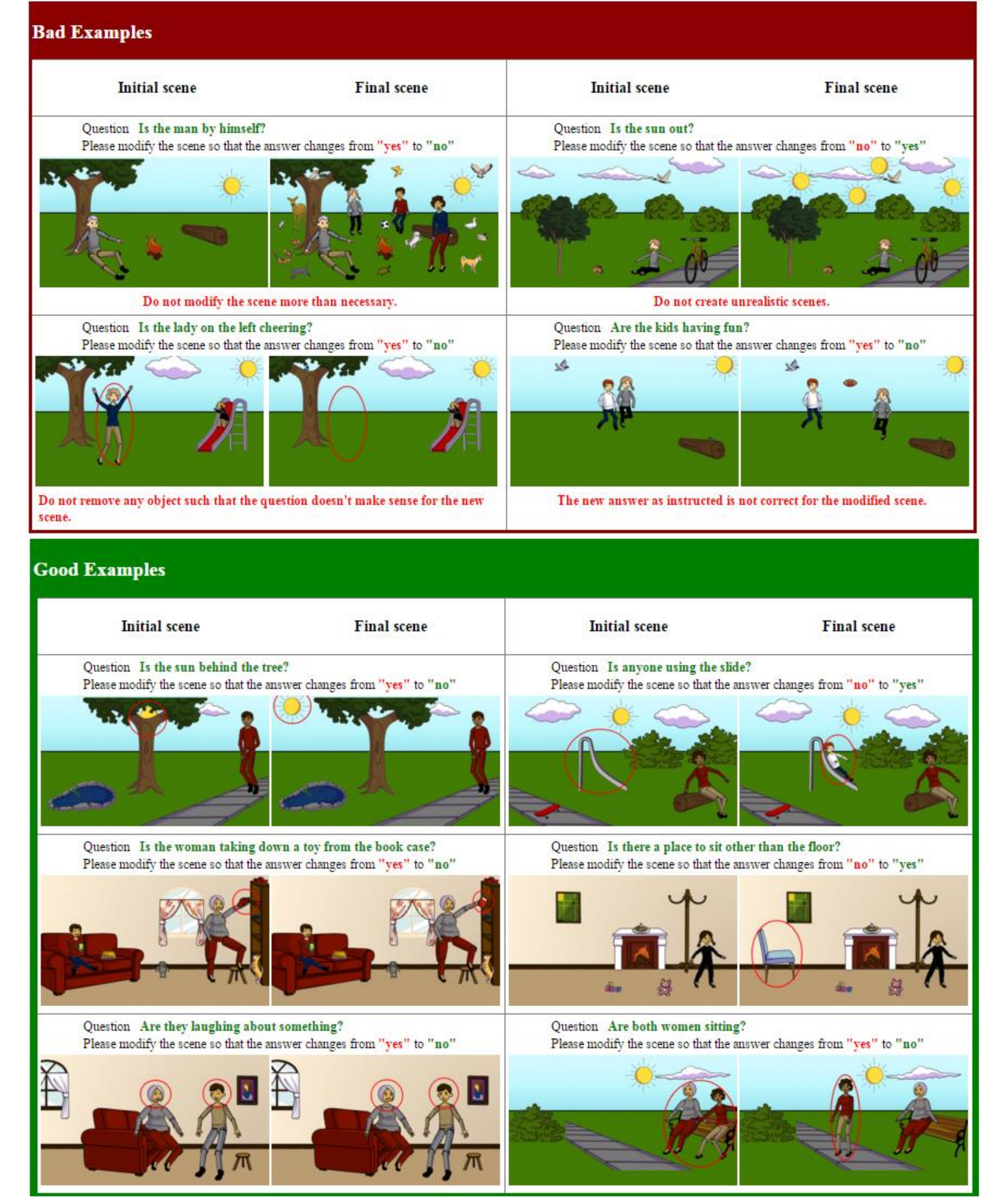}	
   \caption{The good and bad examples shown to AMT workers while collecting complementary scenes.}
\label{fig:interface_examples}
\end{figure*}


\begin{figure*}
\centering
   \includegraphics[width=\textwidth]{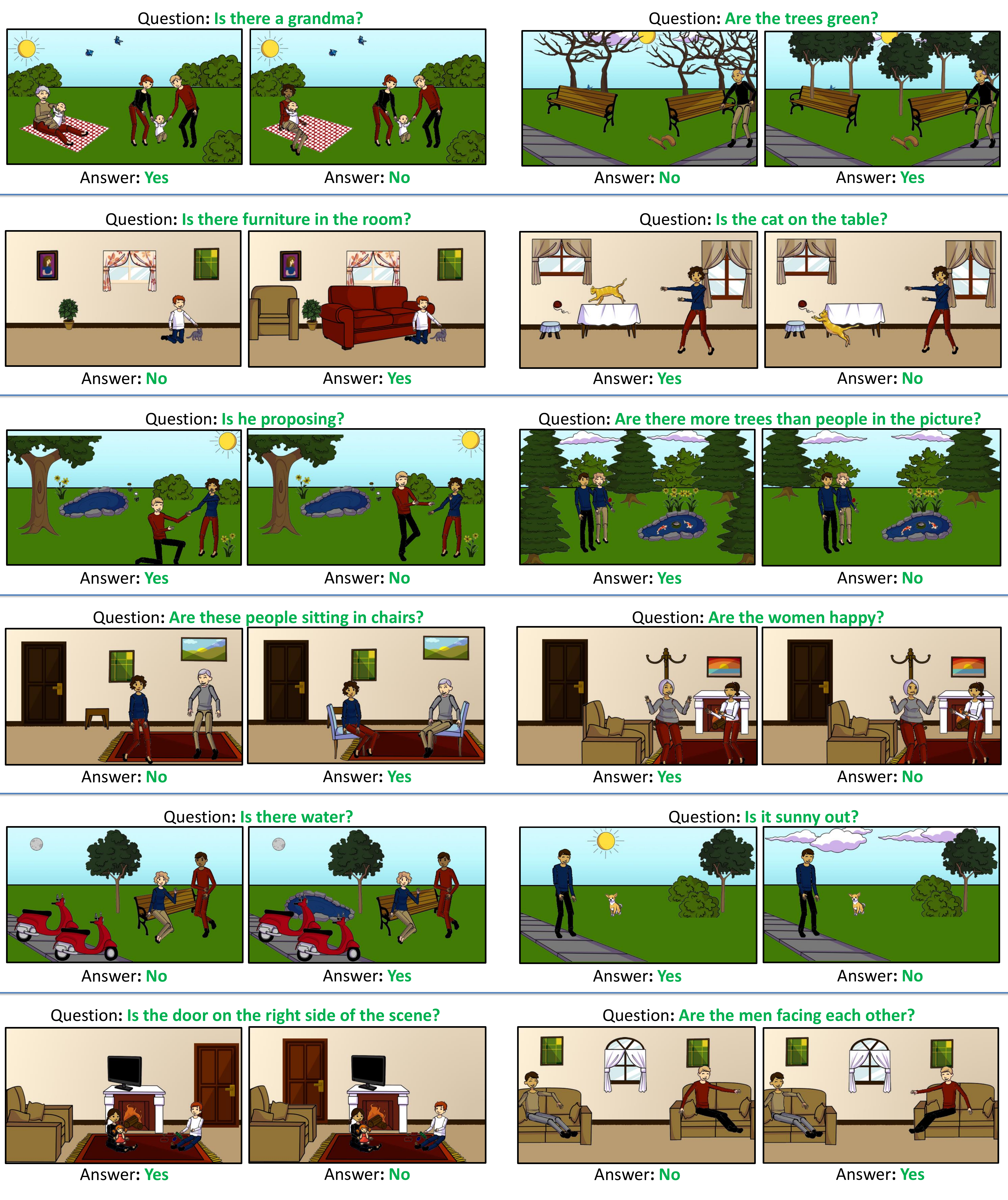}	
   \caption{Example complementary scenes from our balanced dataset. For each pair, the left scene is from VQA dataset \cite{VQA}, and the right scene is the modified version created by AMT workers to flip the answer to the given question.}
\label{fig:dataset}
\end{figure*}
In this section, we analyze the language priors present in the (unbalanced) abstract VQA dataset \cite{VQA}. 
To do this, we implemented an n-gram baseline model to predict answers. 
In our implementation, we used n = 4\footnote{If question has less than 4 words, we use n = length of question.}.
During training, the model extracts all n-grams that start the questions in the training set, and remembers the most common answer for each n-gram.
At test time, the model extracts an n-gram from the beginning of each test question, and predicts the most common answer corresponding to the extracted n-gram for that question.
In case, the extracted n-gram is not present in the model's n-gram memory, the model predicts the answer ``yes'' (which has a higher prior than ``no'').

We found that this model is able to achieve a VQA accuracy of 75.11\%, which is substantially high as compared to PRIOR (``Yes'') baseline accuracy of 68.67\%. 
Qualitatively, the most frequently occurring n-gram in 
our balanced test set,
``is the little girl'' (135 out of total 12,321 questions) has an accuracy of 82.59\% with the answer ``yes''. 
Similarly, the n-grams ``is there a fire'' (61 questions) and ``is the young man'' (35 questions) have accuracies of 97.21\% and 91.28\% respectively with the answer ``yes''. 
In some cases, the bias is towards answering ``no''. 
For instance, the n-gram ``are there leaves on'' has an accuracy of 98.18\% by predicting ``no'' for all 22 questions it occurs in. 
This example clearly demonstrates that humans tend to ask about leaves on trees only when the trees in the images do not have them. There are many more images which contain trees full of leaves, but humans don't ask such questions for those images.

In some cases, the bias in the dataset is because of the limited clipart library. For instance, the predicted answer ``no'' is always correct for the n-grams ``is the door open?'' and ``is it raining?'' because the clipart library has no open doors, and has no rain. Similarly, the n-gram ``is it daytime?'' gets 100\% accuracy with the answer ``yes''.

We believe that LSTMs, known to perform well at remembering sequences, are able to exploit this dataset bias by remembering the most common answer for such n-grams.
Moreover, language only model (which includes LSTMs) gets even higher accuracy (VQA accuracy of 78.9\%) than our n-gram baseline because they probably learn to remember meaningful words in the question instead of the first n words.

%% file: dataset_ap.tex
\section{Abstract library}
\label{abstract_library}

The abstract library consists of two scene types- indoor (``Living/Dining room'') and outdoor (``Park'') scenes. Each scene type is indicated by a different background as shown in \figref{fig:background}, and consists of clipart objects which fit in the respective setting. For instance, indoor scenes can contain indoor objects such as ``pillow'', ``TV'', ``fireplace'', etc., while outdoor scenes can contain outdoor objects such as ``eagle'', ``bike'', ``football'', etc. All clipart objects are categorized into 4 types -- 

1) ``human'': There are 20 paperdoll human models spanning genders, 3 different races and 5 different ages including 2 babies. Each human model can take any one of the available 8 expressions. The limbs are adjustable to allow continuous pose variations. All human models are present in both scene types.

2) ``animal'': There are 10 animal models in indoor scenes, and 31 in outdoor scenes. To keep the scenes realistic, wild animals such as ``deer'', ``raccoon'', ``eagle'', etc. are not available to create indoor scenes. Some of the animal models have been shown in \figref{fig:clipart_objects}.

3) ``large object'': There are 23 large objects present in indoor scenes (e.g. ``door'', ``fireplace'', etc.), and 17 large objects in outdoor scenes (e.g. ``tree'', ``cloud'', etc.). A subset of large objects have been shown in \figref{fig:clipart_objects}.

4) ``small object'': There are 40 small objects present in indoor scenes (e.g. ``toy'', ``pillow'', etc.), and 34 small objects in outdoor scenes (e.g. ``pail'', ``flower'', etc.). A subset of small objects have been shown in \figref{fig:clipart_objects}.

\section{Dataset collection}
\label{collection}

Full instructions of our Amazon Mechanical Turk (AMT) interface to collect complementary scenes, can be seen in \figref{fig:instructions}. To make the task more clear, we also show some good and bad examples (\figref{fig:interface_examples}) to AMT workers. Finally, our AMT interface has been shown in \figref{fig:interface}. 

Some of the complementary scenes from our balanced dataset have been shown in \figref{fig:dataset}.

%% file: qualitative.tex
\section{Qualitative results}
\begin{figure*}
\centering
\includegraphics[width = \textwidth]{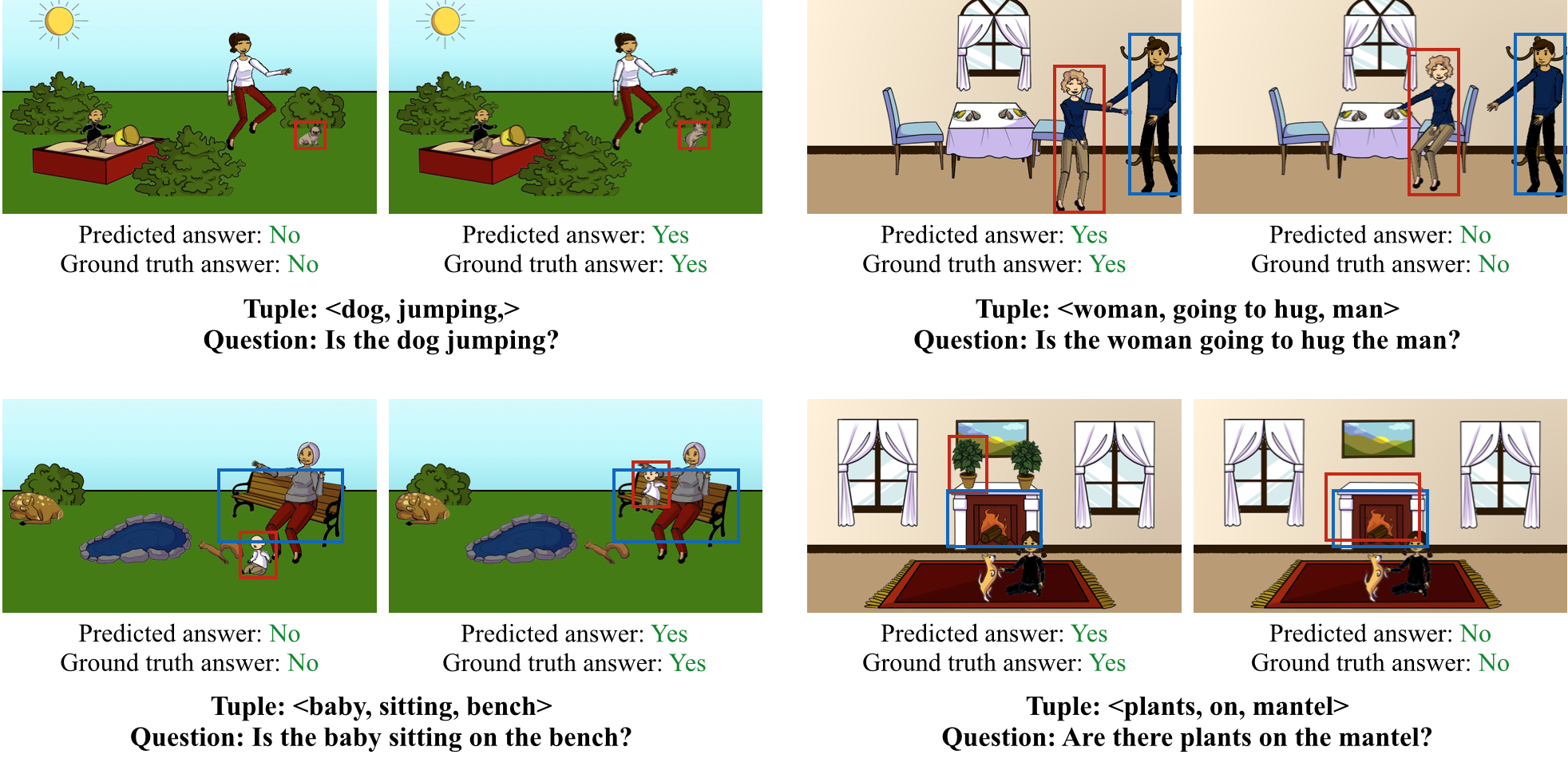}
\includegraphics[width = \textwidth]{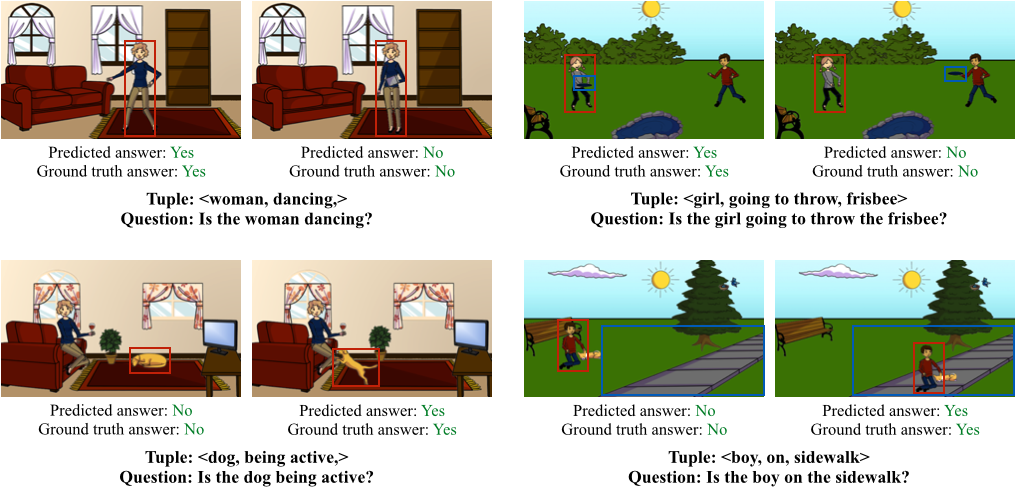}
\includegraphics[width = \textwidth]{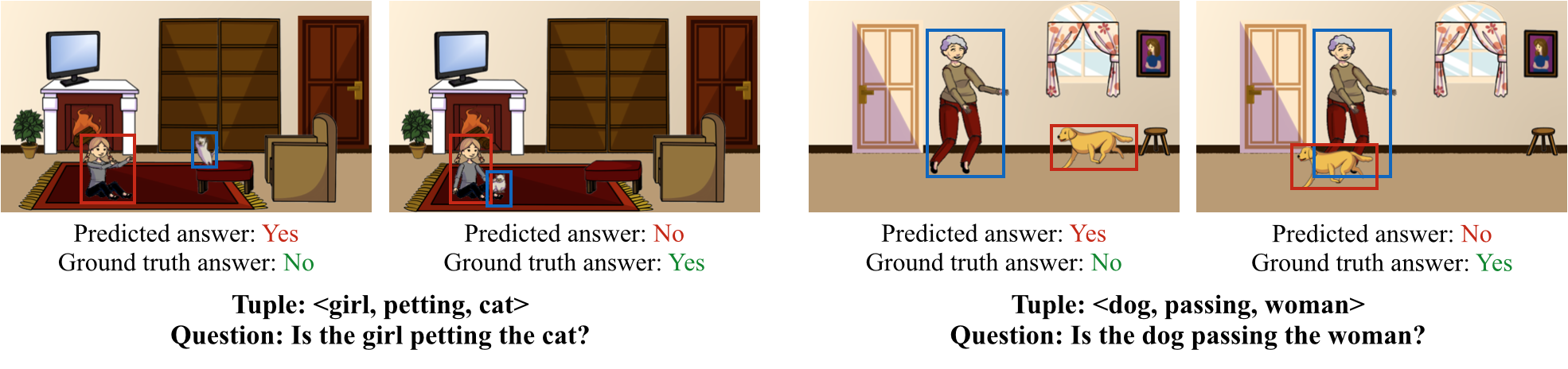}
\caption{Some qualitative results of our approach. The last row shows failure cases. The primary and secondary objects have been shown in red and blue boxes respectively.}
\label{fig:qualitative}
\end{figure*}

We show some qualitative results of our approach in \figref{fig:qualitative}, including some failure cases. In each scene, the primary and secondary objects have been marked in red and blue boxes respectively.

%% file: negation.tex
\section{Issue of Negation}
\label{sec:negation}

Since our approach focuses on the meaningful words in the question, it leads to the issue of poorly handling negative questions. For example, for the questions ``Is the cat on the ground?'' and ``Is the cat not on the ground?'', the extracted tuple would be the same $<$cat, on, ground$>$, but their answers should ideally be opposite. This problem of negation is hard to deal with even in NLP research, but is often ignored because such negative sentences are rarely spoken by humans. For example, in the VQA training dataset, less than 0.1\% of the binary questions contain the word ``not'', ``isn't'', ``aren't'', ``doesn't'', ``don't'', ``didn't'', ``wasn't'', ``weren't'', ``shouldn't'', ``couldn't'', or ``wouldn't''.

%% file: features.tex
\section{Image features}
\label{sec:features_ap}

The image features in our approach are composed of the following 4 parts: 
\begin{compactitem}
\item primary object (P) features (563 dimensions)
\item secondary object (S) features (563 dimensions)
\item relative location features between P and S (48 dimensions)
\item scene-level features (258 dimensions)
\end{compactitem}

P and S features consist of the following parts:
\begin{compactitem}
\item category ID (4 dimensions): category that the object belongs to -- human, animal, large object or small object
\item instance ID (254 dimensions): instance index of the object in the entire clipart library
\item flip attribute (1 dimension): facing left or right
\item absolute locations (50 dimensions): modeled via Gaussian Mixture Model (GMM) with 9 components in 5 different depths separately
\item human features (244 dimensions): composed of age (5 dimensions), gender (2 dimensions), skin color (3 dimensions), pose (224 dimensions), and expressions (10 dimensions)
\item animal features (10 dimensions): pose occurrence for 10 possible discrete poses  
\end{compactitem}

Relative location feature is modeled via another GMM, which is composed of 24 different components. 

Scene level features (258 dimensions) encode object categories and instances of all the objects (other than humans and animals) present in the scene, which consist of 2 parts - category ID and instance ID. 

%% file: tuple_extraction.tex
\section{Details of tuple extraction from raw binary questions}
\label{sec:tuple_extraction_ap}

\subsection{Pre-processing}
\label{sec:preprocess}
We first do the following pre-processing on the raw questions:
\begin{enumerate}
\item We only keep the letters from a to z (both lower cases and upper cases) and digits. 
\item We drop some phrases -- ``do you think'', ``do you guess'', ``can you see'', ``do you see'', ``could you see'', ``in the picture'', ``in this picture'', ``in this image'', ``in the image'', ``in the scene'', ``in this scene'', ``does it look like'', ``does this look like'', ``does he look like'', and ``does she look like'' -- to avoid extra non-meaningful semantic relations in questions. 
\item We make all the letters lower case, but capitalize the first letter. Then we add a question mark at the end of each question.
\end{enumerate} 

\subsection{Summarization}
\label{sec:summarization}

As a first step, we parsed the processed question from~\secref{sec:preprocess} using the Stanford parser \cite{chen14}, resulting in each word in the question being assigned a grammatical entity such as nominal subject (``nsubj''), adverb modifier (``admod''), etc.
Then we follow these steps:

\begin{enumerate}
\item We create a list of all the entities that we would like to keep. The full list is: 
entity list = [``nsubj'', ``root'', ``nsubjpass'', ``case'', ``nmod'', ``xcomp'', ``compound'', ``dobj'', ``acl'', ``advmod'', ``ccomp'', ``advcl'', ``nummod'', ``dep'', ``amod'', ``cc'']. 
In cases where there are more than one word that has been assigned to the same entity, we assign the words different names,
for example, ``nsubj'' and ``nsubj-1''.

\item As we discussed in the main paper, the extracted summary usually starts with entity nominal subject (``nsubj'') or passive nominal subject (``nsubjpass''). So from each parsing result, we first check if ``nsubj'' or ``nsubjpass'' exists. We would like the words that have entity ``nsubj'' or ``nsubjpass'' to be nouns or pronouns. Therefore, we run POS tagger on the whole sentence, specifically we use HunposTagger \cite{hunpos}. 

If the word assigned as ``nsubj'' or ``nsubjpass'' is tagged as noun, then we drop all the words before it. Otherwise, we search for nouns or pronouns from nearby words. 

If the word with entity ``nsubj'' (or ``nsubjpass'') is tagged as pronoun, we would only like to keep more meaningful pronouns such as ``he'', ``she'', etc., rather than ``it'', ``this'' etc. 
So, we created a stop word list for pronouns: [``it'', ``this'', ``that'', ``the'']. Therefore, if the word with ``nsubj'' or ``nsubjpass'' entity is tagged as pronoun and is not present in the above stop list, we keep the question fragments from there.

\textbf{Example1:} Given the question ``Are the children having a good time?'', the parser assigns the word ``children'' as ``nsubj'' and Hunpos tagger tags it as a noun. So we drop all the words before it and output ``children having a good time''.

\textbf{Example2:} Given the question ``Is she playing football?'', the word ``she'' is assigned entity ``nsubj'' by Stanford parser, and is tagged as a pronoun by Hunpos tagger. Then we verify its absence from the stop word list for pronouns. Therefore, we drop all words before it, resulting in ``she playing football'' as summary.

\item Cases where there is neither ``nsubj'' nor ``nsubjpass'' in the question, fall into one of the following three cases: 1) starting with entity ``root'',  2) starting with entity ``nmod'', 3) starting with entity ``nummod''. We directly look for the first noun or pronoun (not in the stop word list) and drop words before it.

\textbf{Example:} Given the question ``Are there leaves on the trees?'', there is no ``nsubj'' or ``nsubjpass'' in it. The word ``leaves'' is tagged as ``root'' and it is the first noun. So, we drop all the words before ``leaves'' and output ``leaves on the trees''.

\item Now that we have fragments of the sentence, we check the entity for each word and delete words whose entity is not in the entity list (described in step 1).

\textbf{Example:} Given a binary question: ``Is the girl pushing the cat off the stool?'', the parsing result from Stanford parser is: Is (aux) the (det) girl (nsubj) pushing (root) the (det) cat (dobj) off (case) the (det) stool (nmod)? First, we search for ``nsubj'', and find the word ``girl''. POS tagger tags it as ``PRP'', which means pronoun, so we drop all the words before it. This results in the fragments ``girl pushing the cat off the stool''.
Lastly, we check if there are any words with entities that are not in the entity list and delete the word ``the'' (entity is ``det''). Therefore, we have the extracted summary as ``girl pushing cat off stool''.
\end{enumerate}

\subsection{Tuple extraction}
\label{sec:tuple_ap}

Now that we have the extracted summaries from raw binary questions, we would like to split them into tuples in the form of primary object (P), relation (R) and secondary object (S). We will describe each of them separately.\\

\textbf{Splitting P argument}
\begin{enumerate}
\item From the extracted summary, we first look for words with entity ``nsubj'' or ``nsubjpass''. If it exists, then we split all the words from the beginning until it as P.

\textbf{Example:} If the extracted summary is ``girl pushing cat off stool'', the word ``girl'' has entity ``nsubj'', so we make P = ``girl''.

\item In some cases, we would like to have a noun phrase instead of one noun word as P. For example, if the summary is ``lady on couch'', we would like to have ``lady'' as P, while for ``lady on couch petting dog'', we would like to split ``lady on couch'' as P. 
In the later case, the summary has the common structure: subject word (e.g. lady) + preposition word (e.g. on) + noun word (e.g. couch).
Usually, noun words in two categories can be used: ``real object'' category, which refer to objects in clipart library, for example, desk, chair etc., or ``location'' category, for example, middle, right, etc. Therefore, we created two such lists, as shown below:

Real object list = [``toy'', ``bird'', ``cup'',
 ``scooter'', ``bench'', ``bush'', ``bike', ``dining chair'', ``plate'', ``bluejay', ``cat', ``blanket'', ``dollhouse'', ``yarn'', ``watermelon'', ``pillow'', ``bread'', ``bat'', ``monkey bars'', ``slide'', ``pet bed'', ``stool'', ``frog'', ``seesaw'', ``sandwich'', ``tape'', ``finch'', ``picture'', ``flower'', ``door'', ``sun'', ``rug'', ``moon'', ``campfire', ``rabbit'', ``utensil'', ``sofa'', ``corn'', ``chair'', ``baseball'', ``butterfly'',  ``sidewalk'', ``turtle'', ``steak'', ``doll'', ``coat rack'', ``mouse'', ``ribs'', ``skateboard'', ``end table'', ``paper'', ``rat'', ``koi'', ``cheese'', ``shovel'', ``camera'', ``apple'', ``marshmallow'', ``pigeon'', ``book'', ``lilypad'', ``cloud'', ``log'', ``stapler', ``notebook'',``bookshelf'', ``dog'', ``hawk'', ``fireplace'', ``raccoon'', ``footstool'', ``mushroom'', ``pie'', ``building toy'', ``tea set'', ``bottle'', ``duck'', ``grill'', ``soccer'', ``tree'', ``pen'', ``cd'', ``game system'', ``scissors'', ``lily pad''
``hamburger'', ``puppy'', ``couch'', ``pond'', ``window'', ``eagle'', ``plant'', ``squirrel'', ``tv'', ``dining table'', ``desk'', ``robin'', ``frisbee'', ``pail'', ``pencil'', ``nest'', ``football'', ``kitten'', ``bee'', ``owl'', ``bone'', ``chipmunk'', ``deer'', ``tongs'', ``beehive'', ``sandbox'', ``bottle'', `` basket'', ``table'', ``bed'', ``bar'', ``pad'', ``shelf'', ``house'', ``ground'', ``cartoon'', ``rope'', ``footstool'']

location list = [ ``left'', ``right'', ``center'', ``top'', ``front'', ``middle'', ``back'']


If we find a word tagged as ``nsubj'' or ``nsubjpass'' by Stanford parser, and is tagged as noun by POS tagger, we check if the next word has entity ``IN'' (meaning proposition word), and the next (or two) word(s) after that are belonging to either the object word list or location word list. Lastly, we check if this is the end of the summary, if not, then we group all the words till here as P, otherwise, we only assign ``nsubj'' or ``nsubjpass'' as P.

\textbf{Example:}, If the extracted summary is ``lady on couch petting dog'', the entity ``nsubj'' corresponds to the word ``lady'', and the next word ``on'' is tagged as ``IN'', followed by the word ``couch'', which is in the object list. Moreover, we check that the word ``couch'' is not the last word of the summary. So we assign ``lady on couch'' to P. 

\item If we could not find ``nsubj'' or ``nsubjpass'' in the summary, we directly look for nouns in the summary. And we assign all the consecutive nouns to P. 

\textbf{Example:} For the question ``Is it night time?'', the extracted summary is ``night time'', in which both the words are nouns. So we assign both the words to P.
\end{enumerate}

\textbf{Splitting S argument}
\begin{enumerate}

\item We drop the words included in P from the extracted summary, and look for nouns in the remaining summary. Once we locate a noun, all the words after it are kept as S.

\textbf{Example:} If the extracted summary is ``lady on couch petting dog'', we assign ``lady on couch'' to P, so we drop them from the summary and thus, we are left with ``petting dog'' from the summary. ``petting'' is not a noun, so we move to the next word. And ``dog'' is a noun. We stop here and make everything after it as S, which in this case is only ``dog''. So S = ``dog''.

\item In some cases, there are some words which modify the nouns, for example, ``pretty girl'', ``beautiful flowers'', etc.. In such cases, we would like the adjectives to be included with the nouns in S. To do so, we look for adjectives, which are tagged by POS tagger, and if there is a noun after this, we keep both of them as S. Otherwise, we only keep the nouns as S.

\textbf{Example:} If the summary is ``scooters facing opposite directions'', P = ``scooters'', so we drop it from the summary, this leaves us with ``facing opposite directions''. ``opposite'' is an adjective and after it, ``directions'' is a noun. SO we keep ``opposite directions'' as S.

\item A minor special case is the occurrence of the phrases -- ``in front of'' and ``having fun with''. These are two phrases that we qualitatively noticed cause issues. In these cases, ``front'' and ``fun'' are tagged as nouns, which confuses our system. So if we detect ``in front of'' or ``having fun with'', we skip them and move to the next word.
\end{enumerate}

\textbf{Splitting R argument}

Since we have already split P and S, we simply assign everything else left in the summary as R. 

\textbf{Example:} If the extracted summary is ``lady on couch petting dog'', P = ``lady on couch'' and S = ``dog'', thus we have R = ``petting''.

%% file: append.tex
\section{Definition of some Stanford typed entities }
We used the Stanford parser in this work. To have better understanding of \secref{sec:summarization}, we list some of the Stanford parser entities here for reference.

\textbf{nsubj: nominal subject} A nominal subject is a noun phrase which is the syntactic subject of a clause. 

\textbf{nsubjpass: passive nominal subject} A passive nominal subject is a noun phrase which is the syntactic subject of a passive clause.

\textbf{dobj: direct object} The direct object of a verb phrase is the noun phrase which is the (accusative) object of the verb.

\textbf{root: root} The root grammatical relation points to the root of the sentence.

\textbf{xcomp: open clausal complement} An open clausal complement (xcomp) of a verb or an adjective is a predicative or clausal complement without its own subject. 

\textbf{advmod: adverb modifier} An adverb modifier of a word is a (non-clausal) adverb or adverb-headed phrase that serves to modify the meaning of the word.

\textbf{ccomp: clausal complement} A clausal complement of a verb or adjective is a dependent clause with an internal subject which functions like an object of the verb, or adjective. 

\textbf{advcl: adverbial clause modifier} An adverbial clause modifier of a verb phrase or sentence is a clause modifying the verb (temporal clause, consequence, conditional clause, purpose clause, etc.).

\textbf{dep: depend} A dependency is labeled as ``dep'' when the system is unable to determine a more precise dependency relation between two words.

\textbf{amod: adjectival modifier} An adjectival modifier of an noun phrase is any adjectival phrase that serves to modify the meaning of the NP

\textbf{cc: coordination} A coordination is the relation between an element of a conjunct and the coordinating conjunction word of the conjunct.